\documentclass{article}
\usepackage{spconf,amsmath,graphicx}
\usepackage{hyperref}
\usepackage[english]{babel}
\usepackage[utf8]{inputenc}
\usepackage{algorithm}
\usepackage{caption}
\usepackage[noend]{algpseudocode}
\usepackage{subfig}
\usepackage{array}
\usepackage{booktabs}
\usepackage{mathtools}
\usepackage{commath}
\usepackage[framemethod=tikz]{mdframed}
\usepackage{wrapfig}

\usepackage{amsmath,amssymb,amsthm}
\usepackage{bm}
\newcommand{\bb}[1]{\bm{\mathrm{#1}}}

\hypersetup{
    colorlinks=true,
    linkcolor=black,
    filecolor=magenta,      
    urlcolor=blue,
    citecolor=black
}

\title{Beholder-GAN: Generation and Beautification of Facial Images with Conditioning on Their Beauty Level}
\makeatletter
\newcommand{\printfnsymbol}[1]{%
 \textsuperscript{\@fnsymbol{#1}}%
}
\makeatother

\name{Nir Diamant\printfnsymbol{1}, Dean Zadok\printfnsymbol{1}, Chaim Baskin, Eli Schwartz, Alex M. Bronstein\thanks{* The authors contributed equally to this work.}}
\address{Computer Science Department, Technion - IIT, Israel}

\begin{document}
%
\maketitle
%


\begin{figure*}[h]
\centering
\subfloat{\includegraphics[width = 0.13\textwidth]{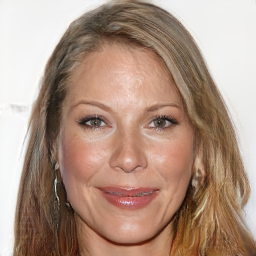}}
\subfloat{\includegraphics[width = 0.13\textwidth]{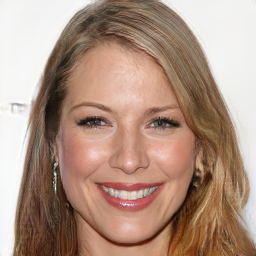}}
\subfloat{\includegraphics[width = 0.13\textwidth]{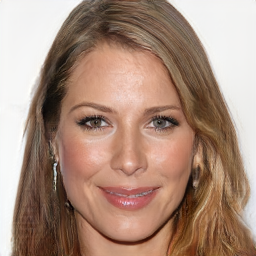}}
\subfloat{\includegraphics[width = 0.13\textwidth]{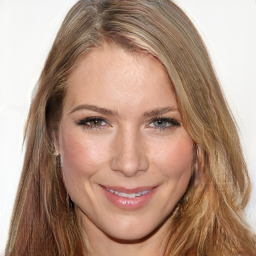}}
\subfloat{\includegraphics[width = 0.13\textwidth]{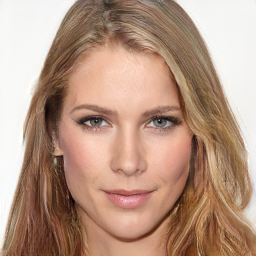}} \\ [-2.35ex]
\subfloat{\includegraphics[width = 0.13\textwidth]{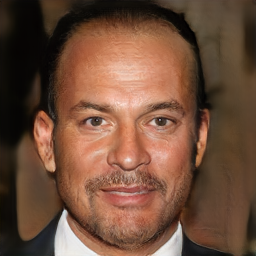}}
\subfloat{\includegraphics[width = 0.13\textwidth]{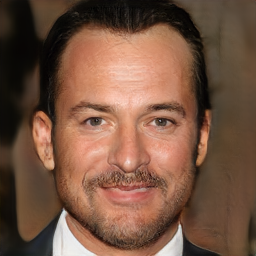}}
\subfloat{\includegraphics[width = 0.13\textwidth]{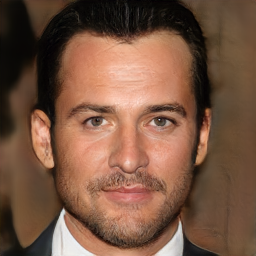}}
\subfloat{\includegraphics[width = 0.13\textwidth]{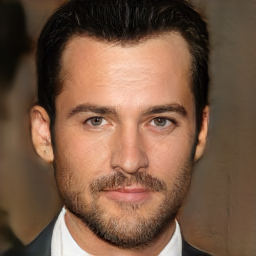}}
\subfloat{\includegraphics[width = 0.13\textwidth]{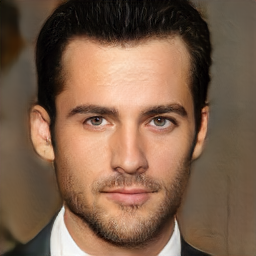}} \\ [-2.35ex]
\subfloat{\includegraphics[width = 0.13\textwidth]{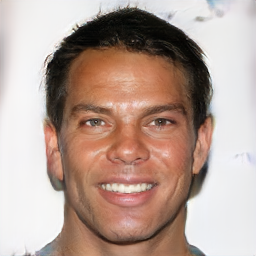}}
\subfloat{\includegraphics[width = 0.13\textwidth]{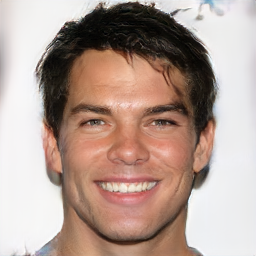}}
\subfloat{\includegraphics[width = 0.13\textwidth]{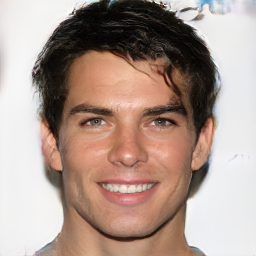}}
\subfloat{\includegraphics[width = 0.13\textwidth]{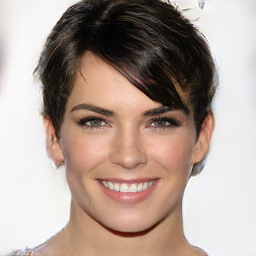}}
\subfloat{\includegraphics[width = 0.13\textwidth]{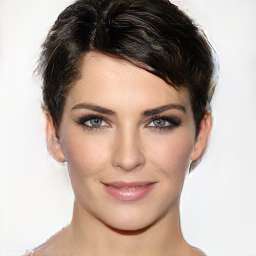}} \\ [-2.35ex]
\subfloat{\includegraphics[width = 0.13\textwidth]{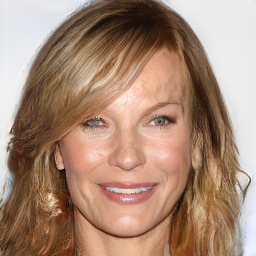}}
\subfloat{\includegraphics[width = 0.13\textwidth]{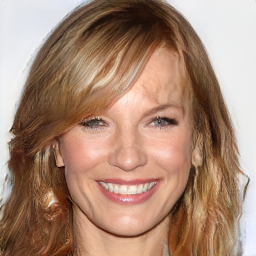}}
\subfloat{\includegraphics[width = 0.13\textwidth]{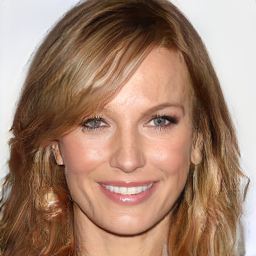}}
\subfloat{\includegraphics[width = 0.13\textwidth]{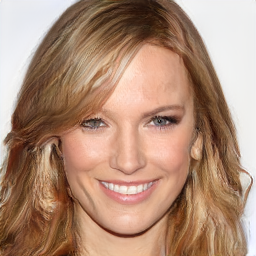}}
\subfloat{\includegraphics[width = 0.13\textwidth]{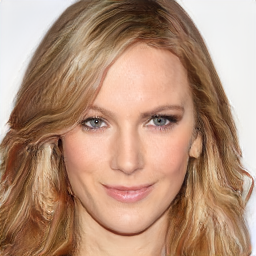}} \\ 
\caption{Generated faces. Each row has the same latent vector but is conditioned on a different beauty score (left to right - least attractive to most attractive). The generated sequences reveal human preferences and biases: for example, older appearances turn into younger ones, masculine faces turn into feminine, and darker skin to brighter. }
\label{figure:collage}
\end{figure*} 

\begin{figure*}[h]
\begin{center}
\setlength\tabcolsep{0pt}
\begin{tabular}{cccccc}
Real & \text{ }\text{ } & \multicolumn{4}{c}{Beautified} \\ 
$\hat{\beta}$ & & $\hat{\beta}+0.1$ & $\hat{\beta}+0.2$ & $\hat{\beta}+0.3$ & $\hat{\beta}+0.4$\\ [-2.1ex]

\subfloat{\includegraphics[width = 0.13\textwidth]{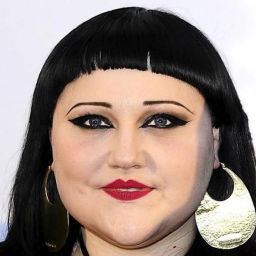}} &  &
\subfloat{\includegraphics[width = 0.13\textwidth]{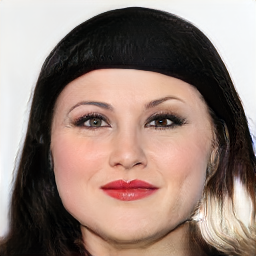}} &
\subfloat{\includegraphics[width = 0.13\textwidth]{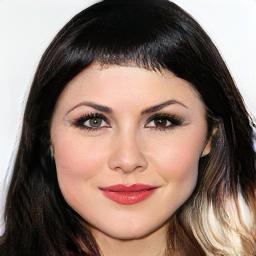}} &
\subfloat{\includegraphics[width = 0.13\textwidth]{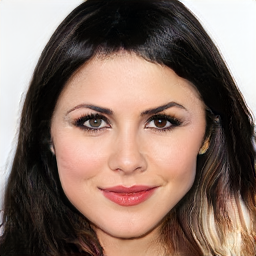}} &
\subfloat{\includegraphics[width = 0.13\textwidth]{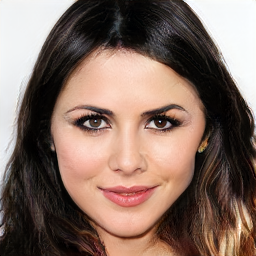}} \\ [-3.1ex]
\subfloat{\includegraphics[width = 0.13\textwidth]{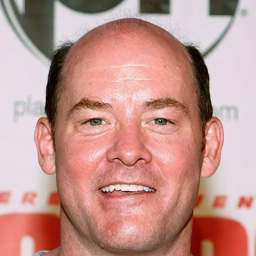}} & &
\subfloat{\includegraphics[width = 0.13\textwidth]{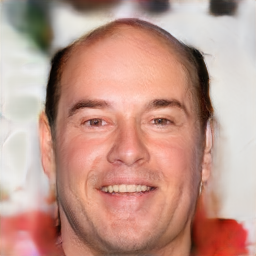}} &
\subfloat{\includegraphics[width = 0.13\textwidth]{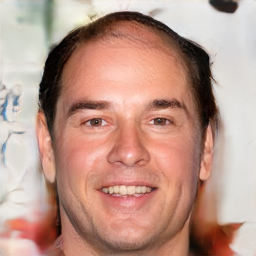}} &
\subfloat{\includegraphics[width = 0.13\textwidth]{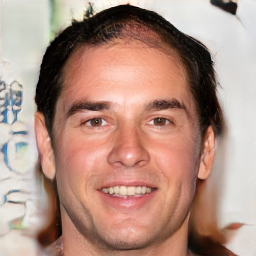}} &
\subfloat{\includegraphics[width = 0.13\textwidth]{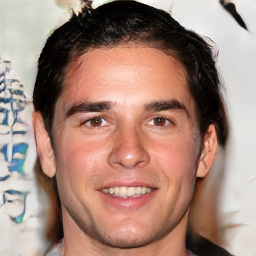}} \\ [-3.1ex]
\subfloat{\includegraphics[width = 0.13\textwidth]{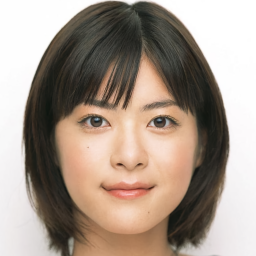}} & &
\subfloat{\includegraphics[width = 0.13\textwidth]{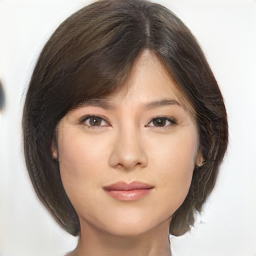}} &
\subfloat{\includegraphics[width = 0.13\textwidth]{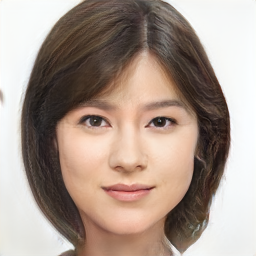}} &
\subfloat{\includegraphics[width = 0.13\textwidth]{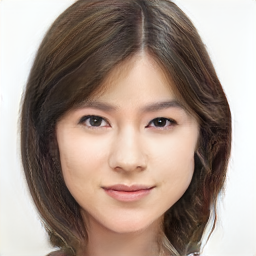}} &
\subfloat{\includegraphics[width = 0.13\textwidth]{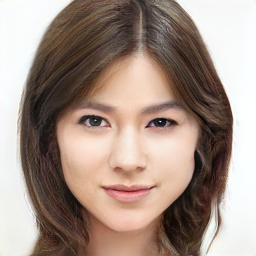}} \\ [-2.9ex]
\end{tabular}
\end{center}
\caption{Beautification of real faces. Left column are the input real faces. To the right are the beautified images with an increasing beauty level ($\hat{\beta}$ is the recovered real face beauty). For $\hat{\beta}+0.1$ we observe reasonable beautification. When further increasing the beauty level it seems that the person identity is not preserved. For privacy and ethical reasons, we refrain from displaying the real faces together with their predicted beauty scores.}
\label{fig:beautification}
\end{figure*}

\begin{abstract}
\emph{"Beauty is in the eye of the beholder."} This maxim, emphasizing the subjectivity of the perception of beauty, has enjoyed a wide consensus since ancient times. In the digital era, data-driven methods have been shown to be able to predict human-assigned beauty scores for facial images. In this work, we augment this ability and train a \emph{generative} model that generates faces conditioned on a requested beauty score. In addition, we show how this trained generator can be used to "beautify" an input face image. By doing so, we achieve an unsupervised beautification model, in the sense that it relies on no ground truth target images. Our implementation is available on: \href{https://github.com/beholdergan/Beholder-GAN}{https://github.com/beholdergan/Beholder-GAN}.

\end{abstract}

\let\thefootnote\relax\footnotetext{\\Corresponding authors:\\Nir Diamant \texttt{nirdiamant21@gmail.com}\\Dean Zadok \texttt{deanzadok@campus.technion.ac.il}}


\section{Introduction}
\label{sec:intro}

Methods for facial beauty prediction and beautification of faces in images have attracted the attention of the computer vision and machine learning communities for a long time \cite{batm,facial-enhancement,deep-face-beautification,facial-beauty-analysis}. The reason goes beyond the importance of these applications, and is probably also related to the inherent challenge in predicting and improving such an utterly subjective attribute as beauty. The fact that beauty is hard to model from first principles makes it a perfect candidate for data-driven methods such as deep learning. 
Over the years, several datasets and methods for facial beauty prediction (FBP) have been suggested, e.g., \cite{dsfbp,scutfbp}. Recently, a new dataset was published \cite{fbp5500} that ranks facial image beauty by group of humans; unlike the previous datasets, the full score distribution for each subject was reported. Our work focuses on the task of generating facial images conditioned on their beauty score. We use it for both generating sequences of images of the same person at different beauty levels, and "beautification" of a given input image.


Generative Adversarial Networks (GANs) are being extensively researched nowadays and were shown to be able to generate realistic high-resolution images from scratch \cite{gans,dcgans}. Nevertheless, the lack of stability in the training process is still noticeable. Implementations such as Unrolled \cite{ugans} and Wasserstein \cite{wgans,iwgan} GANs offer sizeable improvements in stabilizing the training. A recent approach, Progressive Growing of GANs (PGGAN) \cite{pggans} suggested coping with the challenge of generating high-resolution images by learning first through generation of low-resolution images and progressively growing to higher resolutions. 
Another important aspect of GANs is their ability to generate images with conditioning on some attribute, e.g., a class label. These models are often referred to as Conditional GANs (CGANs) \cite{cgans}.

Conditioning vectors can be formed in different structures. One way is the discrete approach where the images are divided into separate classes and fed to the model as a one-hot vector. This method has been used to generate classified images \cite{csgans} or to illustrate face aging \cite{face-aging}. On the other hand, conditioning vectors can be treated as continuous values and fed directly as the input into the model. This method was proposed for synthesizing facial expressions \cite{face-pose-synthesis} or to reconstruct animations based on facial expressions \cite{ganimation}. Regardless of the way the conditioning vector is assembled, usually another output is added to the discriminator where the conditioning vector is predicted and the loss on this output encourages the generated images to belong to the conditional distribution of the the correct class.

In this work, we showcase the ability to generate realistic facial images conditioned on a beauty score using a variant of PGGAN. We use this variant to generate sequences of facial images with the same latent space vector and different beauty levels. This offers insights into what humans consider beautiful and also reveals human biases with regards to age, gender, and race. In addition, we present a method for using the trained generator for recovering the latent vector of a given real face image and use our model to "beautify" it.


\section{Method}
\label{sec:method}

\subsection{A GAN conditioned on a beauty score}
While beauty is a subjective attribute, the scores assigned by different people to facial images tend to correlate. This enabled the creation of datasets of facial images together with their human-labeled beauty scores \cite{geometric-fbp,ava-dataset,assessing-fp}. While the notion of beauty is hard to model mathematically, data-driven methods trained on these datasets are able to predict the beauty scores with remarkable accuracy \cite{humanlike-fbp,transferring-prediction,online-matching}. Another interesting task, which has not been attempted before, is learning a generative model $G$ conditioned on the beauty score:
\begin{equation}
\bb{x} = G(\bb{z}|\beta),
\end{equation}
where $\beta \in [0,1]$ denotes the beauty score, $\bb{z} \sim \mathcal{N}(\bb{0},\bb{I})$ is a random Gaussian vector in some latent space, and $\bb{x}$ is the generated face image.
To ensure that generated image $\bb{x}$ indeed corresponds to the correct beauty level, we also let the discriminator $D$ predict the beauty level and not just the usual real vs. fake probability,
\begin{equation}
(\hat{\mathrm{P}}(\mathrm{real}), \hat{\beta}) = D(\bb{x}),
\end{equation}
and apply an appropriate loss on the beauty score output.
We use the continuous score $\beta$ as input to $G$, and apply the $\ell_2$ loss $(\beta-\hat{\beta})^2$ on the score estimated by the discriminator $D$. In addition, since the beauty score distribution of a single face by multiple beholders can be informative, we input not a single score but a vector of all the ratings available for the face. In Section \ref{sec:experiments} we evaluate these different design choices. With the exception of the the addition of input and output beauty scores, we adopted the architecture and training procedure described in \cite{pggans}.

After training, we can use the trained generator with some fixed $\bb{z}$ and vary the beauty score input $\beta$ to generate faces belonging to the same person but having different beauty levels.










\subsection{Beautification}
\label{ssec:beautification}

A challenge involved in learning beautification of faces is that it must be performed in an unsupervised manner, as we do not have pairs of more and less beautiful images of the same person as would be required for supervised learning. One possible approach for unsupervised learning of transforming images between two domains are methods similar to CycleGAN and image-to-image translation \cite{cyclegan,unitgan}; or the extension to the multi-class case, such as hair color and facial expressions, presented in StarGAN \cite{stargan}. These methods, however, are tailored to the discrete class case and it is not obvious how to adjust them to the case of a continuous attribute such as a beauty score. We propose a method for the beautification of an input facial image using the generator trained, as previously described, in an unsupervised manner -- in the sense that no target image is used to compute the loss.


Given an image $\bb{x}$ and the pre-trained generator $G$, we want to recover the corresponding latent vector $\bb{z}$ and beauty score $\beta$. We do this by initializing with a random $\bb{z}_0$ and $\beta_0$ and performing gradient descent iterations on an aggregate of the $\ell_2$ and VGG losses of the output image compared to the input image. We use a VGG network pre-trained for face recognition and exploit it as feature extractor by removing its last layer. The resulting gradient descent step assumes the form
\begin{eqnarray}
\begin{split}
\bb{z}_{i+1} = & \bb{z}_i - \eta \nabla_{\bb{z}_i} \left(\alpha \norm {G(\bb{z}_i|\beta)-\bb{x}}_2^2 + \right. \\
& \left. (1-\alpha) \norm {\mathrm{VGG}(G(\bb{z}_i\beta_i))-\mathrm{VGG}(\bb{x})}_2^2 \right),
\end{split} \\
\begin{split}
\beta_{i+1} = & \beta_i - \eta \nabla_{\beta_i} \left(\alpha \norm {G(\bb{z}_i|\beta_i)-\bb{x}}_2^2 + \right. \\
& \left. (1-\alpha) \norm {\mathrm{VGG}(G(\bb{z}_i\beta))-\mathrm{VGG}(\bb{x})}_2^2 \right),
\end{split}
\end{eqnarray}
where $\eta$ is the step size, and $\alpha$ governs the relative importances of the $\ell_2$ loss.  
After recovering the latent vector $\hat{\bb{z}}$ encoding to the input face and $\hat{\beta}$, we use the feed forward model $G(\hat{\bb{z}} | \beta_+)$, where $\beta_+$ is an higher beauty level ($\beta_+>\hat{\beta}$), to obtain a similar but more beautiful face.


\subsection{Semi-supervised training}
\label{ssec:method:semi}
Facial images with labeled beauty scores are scarce and, where they exist, do not contain enough variety for training a GAN. To overcome this limitation, we use a semi-supervised approach wherein a model is trained to predict the beauty score of faces based on a limited dataset. The trained model is then used to rate more images, thus creating a richer dataset. Since we condition the GAN on the distribution of scores and not on a single score, we train one predictive model per human rater, e.g., for the SCUT-FBP5500 dataset \cite{fbp5500} with $60$ distinct human raters, $60$ models were trained to predict the scores assigned by each of them.



\begin{table*}[t]
\centering
\begin{tabular}{*7c}
\toprule
Method & \multicolumn{5}{c}{Sliced Wasserstein distance $\times10^3$} & \multicolumn{1}{c}{MS-SSIM} \\
\midrule
{}  & 16x16 & 32x32 & 64x64 & 128x128 & avg. & {} \\
PGGAN \cite{pggans} & \textbf{5.13} & \textbf{2.02} & \textbf{3.04} & \textbf{4.06} & \textbf{3.56} & 0.284 \\
Ours & 8.72 & 4.26 & 6.23 & 11.75 & 7.74 & \textbf{0.274} \\
\bottomrule
\end{tabular}
\caption{Comparison of realism metrics between the unconditioned PGGAN procedure and the proposed conditional GAN}
\label{table:metrics}
\end{table*}

\vspace{-10pt}
\section{Experiments}
\label{sec:experiments}

\subsection{Semi-supervised training}
\label{ssec:semisupervised}

\begin{figure}
\centering
\includegraphics[width = 0.4\textwidth]{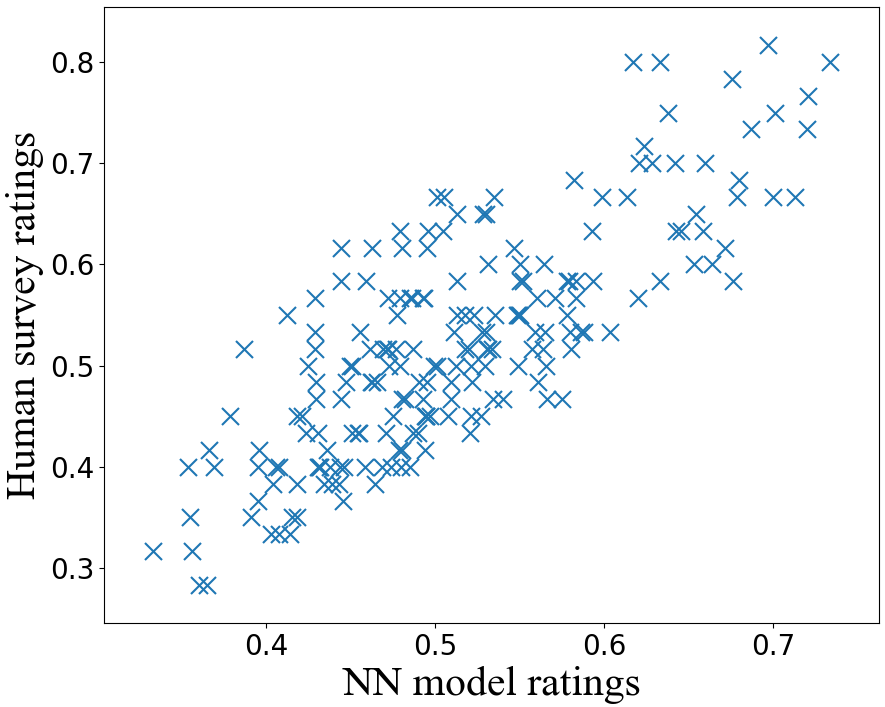}
\caption{Comparing ratings of beauty by humans and a trained neural network model for a new dataset.}
\label{fig:semisup}
\end{figure}

As explained in Section \ref{ssec:method:semi}, we enrich our dataset by training a beauty predictor on one dataset and use it for labeling additional faces. To verify the validity of this idea, we trained a predictive model on the SCUT-FBP5500 dataset \cite{fbp5500} and tested it on $200$ random images from CelebAHQ \cite{pggans}. $60$ VGG models were trained, one per human rater, with the weights initialized from VGG trained on ImageNet. In addition, we ran an online survey where $20$ people (min. age $16$ years; max. age $61$ years; average age $29.5$) rated the beauty score of the $200$ random facial images. The average scores predicted by the trained models compared to the average scores given by human raters are presented in Fig. \ref{fig:semisup}. The correlation between the human  and model ratings is $0.79$, indicating that despite the model ratings being somewhat noisy, they can be used to train our GAN.

\begin{figure}[t]
\begin{center}
\setlength\tabcolsep{0pt}
\renewcommand{\arraystretch}{0}
\begin{tabular}{cc}
\subfloat{\includegraphics[width = 0.118\textwidth]{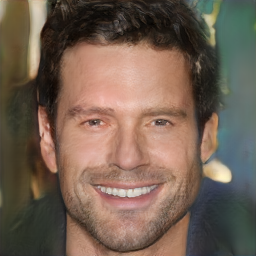}}
\subfloat{\includegraphics[width = 0.118\textwidth]{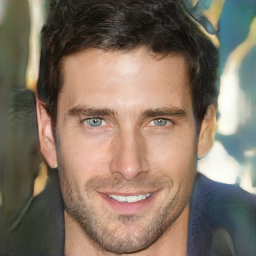}}\hspace{4pt} &
\subfloat{\includegraphics[width = 0.118\textwidth]{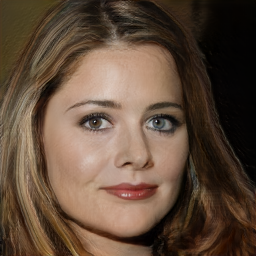}}
\subfloat{\includegraphics[width = 0.118\textwidth]{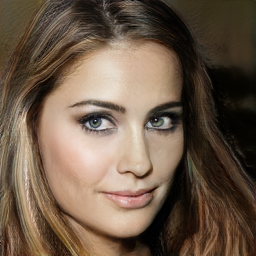}} \\ [-1.5ex]
\subfloat{\includegraphics[width = 0.118\textwidth]{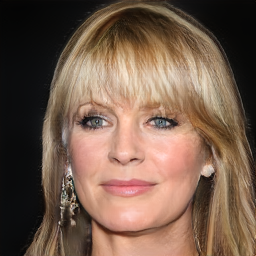}}
\subfloat{\includegraphics[width = 0.118\textwidth]{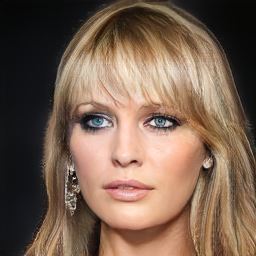}}\hspace{4pt} &
\subfloat{\includegraphics[width = 0.118\textwidth]{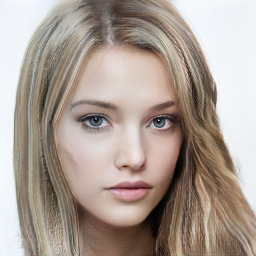}}
\subfloat{\includegraphics[width = 0.118\textwidth]{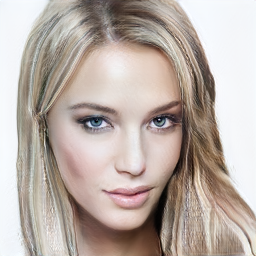}} 
\end{tabular}
\end{center}
\caption{The generative model sense of beauty vs. human rankings. Each pair generated by varying only the generator input beauty score (left lower) and people were asked in a survey to rank them. Only regarding the bottom right pair did the raters and model disagree. 78\% agreement overall.}
\label{fig:pairs_survey}
\end{figure}

\subsection{Face generation and beautification}
\label{ssec:generatingsequences}
We used the previously described methods for labeling the CelebAHQ dataset and trained a GAN for it. We fed each random latent vector $\bb{z}$ to the generator with five beauty scores $\beta \in \{0.1,0.3,0.5,0.7,0.9\}$ to generate five images of supposedly the same person with a different level of beauty. A few examples of the generated sequences are presented in Fig. \ref{figure:collage}. Our evaluation of the results is based on two criteria: the level of realism in the generated images, and the causality between the input beauty score and the generated faces.

To verify that our model generates realistic faces, regardless of the conditioned beauty features, i.e., for quantitative evaluation, we used the Sliced Wasserstein Distance and the MS-SSIM metric employed in the PGGAN evaluation \cite{pggans}. Table \ref{table:metrics} presents the comparison of our conditional GAN with the unconditional PGGAN. While adding the conditioning results in a degradataion in the Sliced Wasserstein Distance, the MS-SSIM metric actually improved.

To evaluate the causality between the input beauty score and the generated faces, we conducted an online survey with pairs from the generated sequences. Each pair consisted of two images with the same latent vector $\bb{z}$ and a distance of $0.2$ in their beauty score $\beta$. The pair was presented to the human raters in a random order; the raters were asked to evaluate which face in the pair appeared more beautiful. A total of $400$ ratings were collected. The percentage of agreement of the human raters with the conditioning score was $78\%$. Fig. \ref{fig:pairs_survey} shows a few examples where the raters agreed or disagreed with the generated beauty score.







We used the generator trained for random face generation for the beautification process as described in Section \ref{ssec:beautification}. Fig. \ref{fig:beautification} presents real faces with the corresponding outputs of the beautification process. The beautified faces are generated by recovering the $\hat{\bb{z}}$ and $\hat{\beta}$ of the real face with gradient descent. Then the feed forward generator is used with the same $\hat{\bb{z}}$ and $\beta \in \{\hat{\beta}+0.1,\hat{\beta}+0.2,\hat{\beta}+0.3,\hat{\beta}+0.4\}$. For real life applications probably only $\hat{\beta}+0.1$ is relevant, as it somewhat preserves the identity of the original face. 

\vspace{-10pt}
\section{Discussion}
\label{sec:discussion}
\vspace{-10pt}

We have shown that despite the subjective nature of beauty, a generative model can learn to capture its essence and generate faces with different beauty levels. We expected a disentanglement between beauty level and the person's identity defined by attributes such as race or gender. In practice, we actually found that when generating two faces with the same latent vector and for enough beauty scores, these attributes tend to change. It might be tempting to call it a "racist algorithm", but we believe it just reflects the subjective, possibly unconscious, biases of the human annotators. We also presented a method to use the trained generative model for the beautification of faces. It should, however, be used with care: While a small increase in the beauty score looks like retouching, a big increase transforms the face into another person.
\section*{Acknowledgments}
The research was funded by ERC StG RAPID.
\bibliographystyle{IEEEbib}
\bibliography{main}

\end{document}